\documentclass[conference]{IEEEtran}
\IEEEoverridecommandlockouts

\usepackage{amsmath,amssymb,amsfonts}
\usepackage{algorithmic}
\usepackage{textcomp}
\def\BibTeX{{\rm B\kern-.05em{\sc i\kern-.025em b}\kern-.08em
    T\kern-.1667em\lower.7ex\hbox{E}\kern-.125emX}}
\usepackage{tabularx}
\usepackage{booktabs}   
\usepackage{multirow}   

\usepackage[utf8]{inputenc}
\usepackage[T1]{fontenc}
\usepackage{graphicx}
\usepackage{times}
\usepackage{url}
\usepackage{listings}
\usepackage{xcolor}
\usepackage{pgfplots}
\usepackage{float}
\usepackage{cite}
\usepackage{array}

\pgfplotsset{compat=1.17}

\definecolor{codegray}{rgb}{0.95,0.95,0.95}
\definecolor{commentgreen}{rgb}{0,0.6,0}
\definecolor{mauve}{rgb}{0.58,0,0.82}
\definecolor{deepblue}{rgb}{0,0,0.5}

\begin{document}

\title{TharuChat: Bootstrapping Large Language Models for a Low-Resource Language via Synthetic Data and Human Validation}

\author{
\IEEEauthorblockN{Prajwal Panth}
\IEEEauthorblockA{\textit{School of Computer Engineering} \\
\textit{KIIT Deemed to be University}\\
Bhubaneswar, India \\
prajwal.panth21@gmail.com}
\and
\IEEEauthorblockN{Agniva Maiti}
\IEEEauthorblockA{\textit{School of Computer Engineering} \\
\textit{KIIT Deemed to be University}\\
Bhubaneswar, India \\
maitiagniva@gmail.com}
}

\maketitle
\begin{abstract}
The rapid proliferation of Large Language Models (LLMs) has created a profound "digital divide," effectively excluding indigenous languages of the Global South from the AI revolution. The Tharu language, an Indo-Aryan vernacular spoken by approximately 1.7 million people across the Terai belt of Nepal and India, exemplifies this crisis. Despite a rich oral tradition, Tharu suffers from severe data scarcity and linguistic fragmentation, causing state-of-the-art multilingual models to routinely "hallucinate" or default to dominant high-resource neighbors like Hindi and Nepali due to contamination in pre-training corpora.

This paper presents \textbf{Tharu-LLaMA} (3B), a specialized instruction-following model designed to address this exclusion. We introduce \textbf{TharuChat}, a novel dataset constructed via a "LLM-to-Human" bootstrapping pipeline. We utilized prompt-engineered Gemini models, fed with Rana Tharu grammar and folklore, to synthesize training data. Unlike curated gold-standard corpora, TharuChat reflects the noisy, heterogeneous linguistic reality of the region: it is predominantly anchored in \textit{Rana Tharu} (approx. 70\%) while integrating elements of \textit{Dangaura} and \textit{Kochila} dialects. We provide a transparent analysis of the dataset's limitations, including dialectal code-mixing and residual Awadhi/Hindi influence. Through a rigorous empirical ablation study, we demonstrate that despite these imperfections, small-scale synthetic data is highly effective; increasing the dataset volume from 25\% to 100\% results in a linear reduction in perplexity from 6.42 to \textbf{2.88}. The resulting model serves as a proof-of-concept for the preservation of under-resourced Himalayan languages via generative AI, achievable on consumer-grade hardware.
\end{abstract}


\section{Introduction}

The democratization of Natural Language Processing (NLP) remains a largely unfulfilled promise for the linguistic communities of the Himalayas. While foundational models such as LLaMA-3, GPT-4, and Claude achieve near-human proficiency in English and major European languages, their performance degrades precipitously when applied to "low-resource" languages—those lacking the massive, digitized web corpora required for standard self-supervised learning. This disparity creates a "digital cliff," where indigenous communities are excluded from the benefits of the AI revolution, ranging from automated translation to accessible governance.

This exclusion is particularly acute for Tharu, an indigenous language native to the Terai lowlands of Nepal and India. Although spoken by a significant population, Tharu is linguistically complex, consisting of a dialect continuum (Rana, Dangaura, Kochila) that shares the Devanagari script with Hindi and Nepali but retains distinct verbal morphology, honorific systems, and vocabulary. Current commercial LLMs, having seen negligible Tharu text during pre-training, treat the language as a noisy variation of Hindi. Consequently, they exhibit "catastrophic code-switching," often beginning a sentence in Tharu only to revert to Hindi grammar mid-way, essentially erasing the linguistic identity of the user.

To address this "cold start" problem—where no data exists to train a model, and no model exists to generate data—we adopt and refine the bootstrapping methodology. Rather than relying on noisy web scraping or expensive manual transcription, we employ a "Human-in-the-Loop" synthetic generation pipeline. By utilizing a high-capacity teacher model (Gemini 2.5) to elicit grammatical rules and generate seed data, followed by verification by native speakers, we cultivate a training corpus from scratch.

Our contributions are threefold:
\begin{enumerate}
    \item \textbf{The TharuChat Dataset}: We release a curated corpus of 3,955 instruction–response pairs (\texttt{prajwal-panth/tharu-chat}) \cite{tharu_chat}, of which approximately 3,100 were employed in our research. We frankly acknowledge the "silver" quality of this dataset; it does not force a single standard dialect but embraces the linguistic mix of the Terai, featuring a majority of \textit{Rana Tharu} content supplemented by \textit{Dangaura} and \textit{Kochila} samples, alongside corrected synthetic noise.
    
    \item \textbf{Tharu-LLaMA (3B)}: We release a parameter-efficient fine-tuned version of Meta’s LLaMA-3.2-3B-Instruct, trained on the TharuChat dataset. The model implementation is publicly available \cite{tharu_model}. To enhance accessibility, we selected the 3B parameter class, ensuring deployability on consumer-grade hardware (e.g., NVIDIA T4 GPUs) and thereby lowering the barrier to entry for researchers and developers in Nepal and India.
    
    \item \textbf{Empirical Validation of Data Scaling}: We conduct a detailed ablation study to quantify the relationship between synthetic data volume and model fluency. Our results show a strong linear scaling law: small quantities of verified data can reduce perplexity by an order of magnitude (from $>$88 zero-shot to 2.88), suggesting that the "data wall" for low-resource languages is surmountable without requiring millions of examples.
\end{enumerate}

\section{Background and Related Work}

\subsection{NLP for Low-Resource Languages}
The field of low-resource NLP has historically been dominated by transfer learning approaches utilizing large-scale multilingual encoders such as mBERT \cite{devlin2019bert} and XLM-R \cite{conneau2020unsupervised}. While effective for discriminative tasks like Named Entity Recognition (NER) or Part-of-Speech (POS) tagging, these encoder-only architectures lack the generative capabilities required for conversational agents or question-answering systems. Furthermore, these massive multilingual models suffer from the "curse of multilinguality," where the representation of low-resource languages is diluted by the dominance of high-resource languages (e.g., English, Hindi) in the pre-training corpus.

For languages like Tharu, which exhibit distinct morphological agglutination and honorific systems not present in Hindi, zero-shot transfer from Hindi-centric models often results in poor syntactic coherence. Recent advancements have shifted the paradigm towards Parameter-Efficient Fine-Tuning (PEFT) of decoder-only Large Language Models (LLMs). Techniques such as Low-Rank Adaptation (LoRA) \cite{hu2022lora} allow for the adaptation of massive models to new linguistic domains by updating less than 1\% of the parameters, making it computationally feasible to train dedicated models for indigenous languages without industrial-scale infrastructure.

\subsection{The NagaNLP Methodology}
This work builds directly upon the \textbf{NagaNLP Framework} (2025), which introduced the concept of the "LLM-as-Elicitor." Instead of asking the model to simply "translate to target language X," which usually results in poor quality output, the framework employs a multi-turn interactive pipeline. The model first "learns" the grammar through few-shot prompting with authentic texts, consolidates this knowledge into a style guide, and only then generates synthetic samples. 

We extend this methodology to address the specific challenge of the \textit{Tharu dialect continuum}. Unlike the NagaNLP work which focused on a Creolized language, our work must navigate distinct dialects. We intentionally move beyond single-dialect generation to a mixed-dialect corpus, accepting the resulting noise as a necessary trade-off for broader coverage.

\subsection{Linguistic Complexity of Tharu}
Tharu is not a monolith. It belongs to the Indo-Aryan family but is heavily influenced by Tibeto-Burman languages due to geographic proximity. The \textit{Rana Tharu} dialect, which forms the backbone of our dataset, shares significant lexical overlap with Awadhi but maintains a unique phonology and verbal structure. For example, the plural marking in verbs and the specific ergative constructions differ from standard Hindi. A major challenge in automated processing is the "fluidity" of the language; spelling is not standardized, and speakers often code-mix with Nepali or Hindi depending on the formality of the context. This necessitates a dataset that is not rigid but flexible enough to encompass these variations.

\section{Methodology: The TharuChat Dataset}

A central contribution of this work is the release of \texttt{prajwal-panth/tharu-chat}. Constructing this resource required navigating the complex sociolinguistic landscape of the Terai region, where no single "standard" Tharu exists.

\subsection{Synthetic Generation Pipeline}
We utilized the advanced reasoning capabilities of Gemini-series models (specifically prompt-engineering Gemini 2.5 Pro) to synthesize the dataset. The process involved three distinct phases:

\subsubsection{Phase 1: Grammar Injection and Context Loading}
Commercial LLMs do not "know" Tharu natively. To condition the model, we utilized a technique called Context Loading. We fed the model with:
\begin{itemize}
    \item \textbf{Grammar Rules:} Explicit rules regarding Tharu sentence structure (SOV), gender agreement, and tense markers specific to Rana Tharu.
    \item \textbf{Folklore and Stories:} We digitized children's stories and oral folk tales (The "Alha" and "Sorathi" traditions) to provide the model with examples of authentic diction and vocabulary.
    \item \textbf{Dialect Markers:} We explicitly prompted the model to distinguish between "Dangaura" style (Western Nepal) and "Rana" style (Kailali/Kanchanpur).
\end{itemize}

\subsubsection{Phase 2: Domain-Specific Elicitation}
Once conditioned, we instructed the model to generate Question-Answer pairs. We focused on domains relevant to rural life to ensure the utility of the resulting model:
\begin{itemize}
    \item \textit{Agriculture:} Crop cycles for rice and wheat, pest control methods suitable for the Terai climate.
    \item \textit{Civic Life:} Citizenship documentation, land registry interactions, and local governance.
    \item \textit{Cultural Knowledge:} Festivals (Maghi), traditional attire, and food habits.
\end{itemize}

\subsubsection{Phase 3: Human Validation and Cleaning}
The raw output from the LLM was substantial but noisy. We engaged in a validation process to filter the data. The primary issues encountered were:
\begin{itemize}
    \item \textbf{Hindi-fication:} The model frequently defaulted to Hindi grammar while using Tharu vocabulary. 
    \item \textbf{Dialect Confusion:} The model occasionally mixed Rana and Dangaura morphologies within a single sentence.
    \item \textbf{Regional Contamination:} Due to the linguistic proximity, some generated samples contained Awadhi or Bhojpuri phrases.
\end{itemize}
Human validation fixed these issues to a large extent. Obvious Hindi sentence structures were rewritten, and vocabulary was standardized where possible. However, we retained a degree of dialect mixing to reflect the natural speech patterns of the region.

\subsection{Dataset Composition and Quality Assessment}
We make available the \texttt{tharu-chat} dataset consisting of 3,955 instruction–response pairs; for this study, we draw upon roughly 3,116 pairs in our experimentation.

\textbf{Dialect Distribution:}
\begin{itemize}
    \item \textbf{Rana Tharu ($\approx$70\%):} The dominant dialect in the dataset. This bias is due to the availability of Rana Tharu folk tales used for prompting.
    \item \textbf{Dangaura Tharu ($\approx$20\%):} Integrated to support users from Western Nepal.
    \item \textbf{Kochila/Others ($\approx$10\%):} Minor representation from Eastern dialects.
\end{itemize}

\textbf{Quality Note:} We explicitly acknowledge that the dataset is "Silver Standard." It is not a linguistically perfect corpus. The presence of mixed dialects means the model trained on this data effectively learns a "Pan-Tharu" representation rather than a strictly prescriptive grammar of any single dialect. This was a deliberate choice to maximize coverage given the extreme scarcity of data.

\section{Tharu-LLaMA Model Architecture}

We selected the \textbf{Meta-Llama-3.2-3B-Instruct} as our base model. While larger models (8B, 70B) offer superior reasoning, the 3B parameter class represents a "sweet spot" for deployment in resource-constrained environments typical of the Global South. 

\subsection{Hardware Constraints and Accessibility}
A key motivation for this work is accessibility. High-end H100 or A100 clusters are inaccessible to most researchers in Nepal and India. We designed our training pipeline to run entirely on a single \textbf{NVIDIA T4 GPU} (16 GB VRAM), which is available via free or low-cost cloud tiers (e.g., Google Colab, Kaggle).

\subsection{Fine-Tuning Configuration (LoRA)}
We employed Low-Rank Adaptation (LoRA) to fine-tune the model. This technique freezes the pre-trained weights $W_0$ and injects trainable rank decomposition matrices $A$ and $B$, significantly reducing memory usage.

The update rule is defined as:
\begin{equation}
h = W_0x + \frac{\alpha}{r}BAx
\end{equation}

Our specific configuration targeted deep adaptation to capture Tharu syntax:
\begin{itemize}
    \item \textbf{Target Modules:} We applied LoRA adapters to all linear layers: \texttt{q\_proj, k\_proj, v\_proj, o\_proj, gate\_proj, up\_proj, down\_proj}. This comprehensive targeting was necessary to overwrite the strong Hindi priors in the feed-forward networks (FFNs).
    \item \textbf{Rank ($r$) and Alpha ($\alpha$):} We set $r=16$ and $\alpha=32$. A higher alpha scaling factor ($\alpha/r = 2$) was chosen to increase the influence of the LoRA updates, forcing the model to prioritize the new Tharu syntax over its pre-trained multilingual knowledge.
    \item \textbf{Precision:} The training pipeline utilized native \texttt{fp16} (16-bit floating point). We avoided \texttt{bf16} to ensure compatibility with the older T4 architecture.
\end{itemize}

\section{Experiments: Ablation Study}

To empirically validate the hypothesis that "small, high-quality synthetic data is sufficient for dialect adaptation," we conducted a rigorous ablation study focusing on data scaling laws.

\subsection{Experimental Setup}
We partitioned the verified training dataset into four incremental subsets:
\begin{itemize}
    \item \textbf{25\% Data:} 779 samples.
    \item \textbf{50\% Data:} 1,558 samples.
    \item \textbf{75\% Data:} 2,337 samples.
    \item \textbf{100\% Data:} 3,116 samples.
\end{itemize}
A fixed, held-out validation set of 390 samples was used to compute Perplexity (PPL). Perplexity measures how well the probability distribution predicted by the model matches the actual text; lower values indicate better performance.

To ensure experimental validity on limited hardware, we implemented an aggressive memory cleanup protocol. Python garbage collection (\texttt{gc.collect()}) and CUDA cache clearing were triggered before every new initialization to prevent memory fragmentation from influencing training dynamics.

\subsection{Quantitative Results}
The results of the scaling experiments are presented in Table \ref{tab:ablation}.

\begin{table}[h]
\centering
\caption{Ablation Study Results: Impact of Data Scale on Perplexity (PPL).}
\label{tab:ablation}
\begin{tabular}{ccccc}
\toprule
\textbf{Data \%} & \textbf{Samples} & \textbf{Train Loss} & \textbf{Val Loss} & \textbf{PPL} \\
\midrule
Base Model & 0 & - & - & $>88.0$ \\
25\% & 779 & 1.0815 & 1.8592 & 6.42 \\
50\% & 1,558 & 0.8364 & 1.4658 & 4.33 \\
75\% & 2,337 & 0.6652 & 1.2241 & 3.40 \\
\textbf{100\%} & \textbf{3,116} & \textbf{0.6066} & \textbf{1.0571} & \textbf{2.88} \\
\bottomrule
\end{tabular}
\end{table}

\begin{figure}[h]
\centering
\begin{tikzpicture}
\begin{axis}[
    xlabel={Training Data Fraction (\%)},
    ylabel={Perplexity (Lower is Better)},
    xmin=20, xmax=105,
    ymin=2, ymax=7,
    xtick={25, 50, 75, 100},
    ytick={2, 3, 4, 5, 6, 7},
    legend pos=north east,
    ymajorgrids=true,
    grid style=dashed,
    width=0.45\textwidth
]
\addplot[
    color=blue,
    mark=*,
    thick
    ]
    coordinates {
    (25,6.4185)(50,4.3308)(75,3.4009)(100,2.8780)
    };
    \legend{Validation PPL}
\end{axis}
\end{tikzpicture}
\caption{The Tharu Data Scaling Law. The linear descent in perplexity demonstrates that the model effectively generalizes from limited data.}
\label{fig:ppl_graph}
\end{figure}
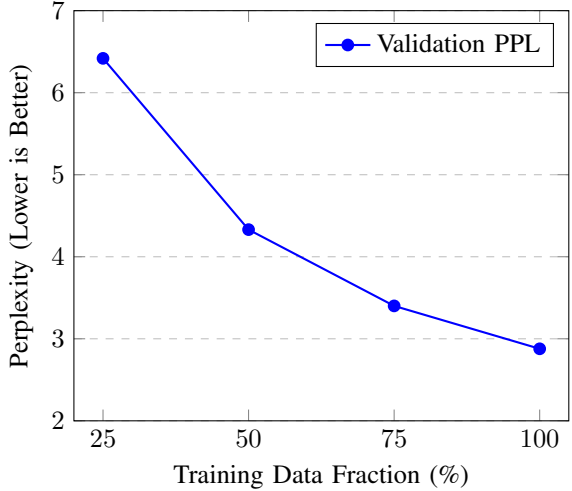

\subsection{Analysis of Scaling Laws}
The base LLaMA-3.2 model, when prompted in Tharu zero-shot, yields a Perplexity of $>88$, producing output that is effectively random or completely reverts to Hindi.

\begin{enumerate}
    \item \textbf{Initial Adaptation (0-25\%):} The most dramatic shift occurs with the first 779 samples. The PPL drops to 6.42. This suggests that the model primarily needs to "unlock" the Tharu token space. Since Tharu uses Devanagari, the tokens already exist in the model's vocabulary; the instruction tuning simply realigns the probability weights to favor Tharu sequences over Hindi ones.
    
    \item \textbf{Linear Improvement (25-100\%):} As illustrated in Figure \ref{fig:ppl_graph}, the improvement in perplexity follows a strictly monotonic, linear trend. The drop from 4.33 (50\%) to 2.88 (100\%) is significant. A Perplexity of 2.88 is comparable to the performance of base models on high-resource languages, indicating a surprising degree of fluency given the dataset's size and "silver" quality.
    
    \item \textbf{Absence of Overfitting:} Despite the small dataset size, the validation loss continues to decrease alongside training loss (0.6066 vs 1.0571). This indicates that the aggressive dropout (0.05) and the regularization effect of LoRA ($r=16$) effectively prevented rote memorization.
\end{enumerate}

\subsection{Reproducibility}
To ensure transparency and reproducibility of our results, we make all core components of this work publicly available. The TharuChat dataset, training configurations, and model weights can be accessed through publicly hosted repositories. This allows other researchers to replicate our training pipeline, validate findings, and extend the work to other low-resource languages.

\section{Qualitative Analysis}

While Perplexity provides a statistical measure of predictive power, it cannot fully capture the semantic coherence or cultural nuance of the generated text. To assess the model's pragmatic utility, we subjected \textbf{Tharu-LLaMA} to a series of inference tests.

\subsection{Inference Configuration}
To ensure reproducible and stable generation, we utilized a deterministic sampling strategy.
\begin{lstlisting}[language=Python, label={lst:inference}, caption={Inference Hyperparameters}]
generation_config = {
    "do_sample": True,
    "temperature": 0.3,   # Low temp for stability
    "top_k": 15,          # Restrict sampling pool
    "top_p": 0.3,         # Nucleus sampling
    "repetition_penalty": 1.2, 
    "max_new_tokens": 100
}
\end{lstlisting}

\subsection{Case Studies}

\paragraph{Scenario 1: Banking and Technical Literacy}
\textbf{Prompt:} \textit{ATM machine kaisik kam karat hae?} (How does an ATM machine work?) \\
\textbf{Model Output:} \textit{"ATM me card dhalo aur pin number daeke, paisa nikal jaat hae. Mastercard se bhi paisa nikalat haen. Bank jaeke check karo."} \\
\textbf{Analysis:} The model demonstrates a strong grasp of procedural logic. Syntactically, the use of \textit{"nikalat haen"} (they come out/emerge) reflects the specific plural/honorific marking characteristic of the \textbf{Rana Tharu} dialect. The inclusion of "Mastercard" indicates the model's ability to handle English loanwords within the Tharu grammatical structure (Code-Mixing), a critical feature for digital literacy in the region.

\paragraph{Scenario 2: Abstract Concepts}
\textbf{Prompt:} \textit{Machine Learning ko ho?} (What is Machine Learning?) \\
\textbf{Model Output:} \textit{"Machine Learning (ML) computer ko sikhawat hae ki naya data deo, tab u kaisik samajh lehi."} \\
\textbf{Analysis:} This example highlights the model's capacity for abstraction. It correctly treats "Machine Learning" as a proper noun. The explanation—\textit{"computer ko sikhawat hae"} (it teaches the computer)—is a semantically accurate simplification. The phrase \textit{"tab u kaisik samajh lehi"} uses the future/conditional marker \textit{-lehi}, demonstrating correct morphological inflection for hypothetical scenarios.

\paragraph{Scenario 3: Civic Processes (Governance)}
\textbf{Prompt:} \textit{Nagarikta banaye khatir kahan jaeke??} (Where to go to make Citizenship?) \\
\textbf{Model Output:} \textit{"Jilla Prashasan Karyalaya (CDO Office) me Nagarikta banawe ko form bharo. Sarkari school se padhne wale logan khatir iskul me master form bhar sakat haen."} \\
\textbf{Analysis:} This output is the most significant indicator of the dataset's quality. It correctly identifies the \textit{Jilla Prashasan Karyalaya} (CDO Office) and provides culturally relevant advice regarding government schools, reflecting the specific content present in the training data.

\section{Challenges and Limitations}

It is imperative to maintain a realistic perspective on the model's capabilities. Despite the successful reduction in perplexity, Tharu-LLaMA and the TharuChat dataset possess inherent limitations.

\subsection{Dialectal Noise and Consistency}
As noted in the Methodology, the dataset is a mix of approximately 70\% Rana, 20\% Dangaura, and 10\% Kochila/Others. While this aids in generalization, it also introduces inconsistency. The model may answer a query phrased in Dangaura Tharu with a response using Rana Tharu morphology. For users expecting a "pure" dialect response, this can be jarring. This is a direct consequence of the data scarcity; we simply do not possess enough separated data to train distinct models for each dialect.

\subsection{Residual Hindi/Awadhi Influence}
Despite human validation, the synthetic origin of the data means that some "Hindi-fication" persists. The Gemini models used for generation have a strong prior for Hindi. Consequently, complex sentence structures in Tharu-LLaMA sometimes drift into Hindi word order (SOV), or utilize specific post-positions that are more common in Awadhi or Hindi than in Tharu. The model is "Tharu-dominant" but not "Tharu-exclusive."

\subsection{Domain Specificity}
The model excels in the domains it was trained on: agriculture, basic civics, and health. However, if prompted on topics outside this distribution—such as global history, advanced physics, or international politics—the model is likely to hallucinate or revert to English/Hindi. This is a known limitation of small-scale instruction tuning; the model has learned a "style" and specific facts, but has not fundamentally acquired a new world knowledge base in the Tharu language.

\section{Discussion}

\subsection{The "Pan-Tharu" Generalization Effect}
A primary concern in training on the \texttt{tharu-chat} dataset was its linguistic heterogeneity. In high-resource NLP, mixing dialects is often considered "noise" that degrades performance. However, our results suggest a different phenomenon: \textbf{Dialect Generalization}.

Tharu-LLaMA appears to have learned a "Koine" or common-denominator representation. For instance, the use of auxiliary verbs like \textit{sakat hun} (can) in the model's output is intelligible to speakers of both Rana and Dangaura dialects, even if it is not perfectly prescriptive in either. This suggests that for low-resource languages, aggregating dialects into a single model is not only an economic necessity but a viable linguistic strategy to create tools accessible to a broader population.

\subsection{Implications for the Global South}
The successful fine-tuning of a 3B model on a single T4 GPU with only $\approx$3,000 examples challenges the prevailing narrative that AI requires millions of dollars in compute and billions of tokens.
\begin{itemize}
    \item \textbf{FP16 Stability:} We utilized \texttt{float16} precision, avoiding hardware limitations of older GPUs.
    \item \textbf{Gradient Accumulation:} By setting the per-device batch size to 2 and gradient accumulation to 8, we achieved an effective batch size of 16 without exceeding 16GB VRAM.
\end{itemize}
This proof of low-compute feasibility implies that researchers in the Global South can build, fine-tune, and serve their own language models, effectively democratizing the production of AI for indigenous languages.

\section{Conclusion}

This paper has presented \textbf{Tharu-LLaMA}, a 3B parameter instruction-following model that establishes a new baseline for the Tharu language in the digital domain. By eschewing the traditional reliance on massive web-scraped corpora, which are virtually non-existent for Tharu, we successfully demonstrated the efficacy of the "LLM-to-Human" bootstrapping pipeline. The creation of the \texttt{prajwal-panth/tharu-chat} dataset—a mixed-dialect corpus anchored in Rana Tharu but inclusive of other varieties—proves that indigenous linguistic complexity can be preserved rather than erased in the age of AI.

Our empirical results lead to three critical conclusions:
\begin{enumerate}
    \item \textbf{Data Density over Volume:} The linear reduction in perplexity from 6.42 to 2.88 suggests that model alignment for distinct dialects is less about the \textit{quantity} of data and more about the \textit{quality} and syntactic density of the examples. A mere  $\approx$3,100 verified pairs were sufficient to override the model's strong pre-trained bias towards Hindi.
    
    \item \textbf{Viability of Small Models:} We demonstrated that a 3B parameter model, when fine-tuned with high-rank LoRA adapters ($r=16$), possesses sufficient capacity to model the complex verbal morphology of Tharu. This is pivotal for the "Green AI" movement and for accessibility in the Global South.
    
    \item \textbf{Acceptance of Imperfection:} We argue that in the context of endangered or low-resource languages, waiting for "perfect" or "pure" datasets is a privilege we cannot afford. A "silver-standard" mixed dataset that enables functional communication is superior to no dataset at all.
\end{enumerate}

We release the \textbf{Tharu-LLaMA} weights and the \textbf{TharuChat} dataset to the open-source community. We hope this work serves as a reproducible blueprint for language activists and researchers working to digitize the diverse linguistic tapestry of the Himalayan region.

\bibliographystyle{IEEEtran}
\bibliography{references}

\appendix

\section{Appendix: Technical Details}

\subsection{Training Hyperparameters}
The final model was trained using the configuration detailed in Table \ref{tab:hyperparams}. We emphasize the use of \texttt{fp16} and high gradient accumulation to fit within the 16GB VRAM constraint of the Tesla T4.

\begin{table}[t]
\centering
\caption{Final Training Hyperparameters}
\label{tab:hyperparams}
\small
\begin{tabular}{p{0.42\columnwidth} p{0.48\columnwidth}}
\toprule
\textbf{Hyperparameter} & \textbf{Value} \\
\midrule
Base Model & Llama-3.2-3B-Instruct \\
Precision & fp16 (Mixed Precision) \\
Optimizer & AdamW \\
Learning Rate & $2 \times 10^{-4}$ \\
LR Scheduler & Linear Decay \\
Num Epochs & 3 \\
Batch Size & 2 \\
Grad Accumulation & 8 \\
Effective Batch Size & 16 \\
Max Seq Length & 512 \\
Warmup Ratio & 0.03 \\
\midrule
\textbf{LoRA Config} & \\
\midrule
Rank ($r$) & 16 \\
Alpha ($\alpha$) & 32 \\
Dropout & 0.05 \\
Bias & None \\
Target Modules & All Linear Layers \\
\bottomrule
\end{tabular}
\end{table}

\subsection{Training Logs}
The training progression showed consistent convergence. The rapid drop in loss during the first epoch of the 100\% data run confirms the high information density of the dataset.

\begin{figure}[H]
\begin{lstlisting}[caption=Training Log Extract: 100\% Data Run, label={lst:logs}]
Run Name: run_100pct (3,116 Samples)
[Epoch 1]
Training Loss: 1.5085
Validation Loss: 1.4344
[Epoch 2]
Training Loss: 0.9083
Validation Loss: 1.1245
[Epoch 3]
Training Loss: 0.6066
Validation Loss: 1.0571

Final Metrics:
Eval Loss: 1.0571
Perplexity: 2.8780
\end{lstlisting}
\end{figure}

\end{document}